\def\year{2021}\relax
\documentclass[letterpaper]{article} % DO NOT CHANGE THIS
\usepackage{aaai21}  % DO NOT CHANGE THIS
\usepackage{times}  % DO NOT CHANGE THIS
\usepackage{helvet} % DO NOT CHANGE THIS
\usepackage{courier}  % DO NOT CHANGE THIS
\usepackage[hyphens]{url}  % DO NOT CHANGE THIS
\usepackage{graphicx} % DO NOT CHANGE THIS
\usepackage{graphicx}  % DO NOT CHANGE THIS
\usepackage{natbib}  % DO NOT CHANGE THIS AND DO NOT ADD ANY OPTIONS TO IT
\usepackage{caption} % DO NOT CHANGE THIS AND DO NOT ADD ANY OPTIONS TO IT
\usepackage{multirow}

\title{Reinforcement learning for the privacy preservation and manipulation of eye tracking data}
\author {
        Wolfgang Fuhl,\textsuperscript{\rm 1}
        Efe Bozkir,\textsuperscript{\rm 1}
        Enkelejda Kasneci, \textsuperscript{\rm 1}\\
}
\affiliations {
   \textsuperscript{\rm 1} University T\"ubingen, Sand 14, 72076 T\"ubingen, Germany \\
    wolfgang.fuhl@uni-tuebingen.de, efe.bozkir@uni-tuebingen.de, enkelejda.kasneci@uni-tuebingen.de
}

\begin{document}

\maketitle

\begin{abstract}
In this paper, we present an approach based on reinforcement learning for eye tracking data manipulation. It is based on two opposing agents, where one tries to classify the data correctly and the second agent looks for patterns in the data, which get manipulated to hide specific information. We show that our approach is successfully applicable to preserve the privacy of the subjects. For this purpose, we evaluate our approach iteratively to showcase the behavior of the reinforcement learning based approach. In addition, we evaluate the importance of temporal, as well as spatial, information of eye tracking data for specific classification goals. In the last part of our evaluation, we apply the procedure to further public data sets without re-training the autoencoder or the data manipulator. The results show that the learned manipulation is generalized and applicable to unseen data as well.
\end{abstract}

\section{Introduction}
Due to the spread of the eye tracking technology over many fields~\cite{WF042019} and its use in everyday life, the specific information content in the eye tracking signal becomes more and more important~\cite{bulling2010toward,majaranta2014eye,032017,0320170,ACTNEURO2017,Bahmani2016,C2019,FFAO2019}. This is mainly due to the fact that the gaze signal consists of very rich information and on the other hand that it cannot be turned off or consciously controlled by humans~\cite{hansen2003command,stellmach2012look,WTDTWE092016,WTDTE022017,WTE032017,WTCDAHKSE122016,WTCDOWE052017,WDTTWE062018,VECETRA2020,CORR2017FuhlW1,ETRA2018FuhlW}. Many applications use this signal, however, still little value is placed on the anonymization of the signal. This is partly due to the fact that the topic of differential privacy has come into the focus of eye tracking research recently~\cite{steil2019privaceye,steil2019privacy,liu2019differential}, but also to the challenge of finding specific patterns in the signal makes a person identifiable. 

Initially why the personal information should be protected in eye and gaze tracking applications along with the person specific patterns contained in the signals including age, gender, personal preference and health was mentioned by~\cite{liebling2014privacy}. This information poses a new challenge to modern eye tracking systems, which is to hide this information. Differential privacy is one approach that achieves privacy of individuals' identities by adding randomly generated noise by keeping privacy-utility ratio acceptable. It usually works in case of prefabricated features; however, modern machine learning techniques such as convolutional neural networks (CNNs) are able to adapt their feature extractors. In addition, differential privacy is vulnerable to temporal correlations in the signals as independently generated noise can be helpful for the adversaries. As eye tracking data is temporally correlated in its nature, differential privacy approaches usually provide less privacy than claimed~\cite{8269219}. Additionally, it would be more interesting to find specific patterns either in the stimulus or, as in this paper, in the scan path, which we can remove from the signal. This insight can be also used in many other areas gaze guidance~\cite{latif2014art,kano2011perceptual} or expertise evaluation~\cite{gegenfurtner2011expertise,kunze2013towards}.

In this paper, we present an approach that is able to learn an image manipulation to hide specific information while preserving utility. Our approach uses reinforcement learning on the sparse representation learned by an autoencoder. This combination allows to manipulate general patterns in an image, since the autoencoder has to reconstruct it based on a reduced set of values. This reduced set can be found in the central part of the autoencoder. It is also called bottleneck, and the following transposed convolutions of the autoencoder reconstruct the image on the basis of the reduced set. Meaning that, those values represent patterns in an image that are manipulated by an agent in our approach. This agent tries to hide specific information by manipulating those values. Another agent tries to train new classifiers to adapt to the manipulated data. The retraining allows our approach to diminish personal patterns in the data since the classifiers adapt to the manipulated data. The main contributions of our work are as follows.
\begin{description}
	\item[1] A novel approach to remove patterns from eye tracking data that contain personal information which achieves a similar goal as differential privacy.
	\item[2] Being independent of static features due to the iterative usage of CNNs.
	\item[3] Identification of general patterns in the data instead of adding randomly generated noise as it is done in differential privacy.
	\item[4] Possibility of specifying the information type that should be hidden in the data.
\end{description}

\section{Related Work}
As we deal with two main topics including privacy using eye movements and reinforcement learning for manipulation of the eye tracking data, we organize this section accordingly.

\subsection{Eye movements and privacy}
The rich information content is available in human eye movements~\cite{FCDGR2020FUHL,fuhl2018simarxiv,ICMIW2019FuhlW1,ICMIW2019FuhlW2,EPIC2018FuhlW,C2019,FFAO2019,NNETRA2020} and it has been shown in several studies. Cognitive load~\cite{matthews1991pupillary}, attention and personal interest in the scene~\cite{hess1960pupil} can be extracted using pupil dilation. Mental disorders such as Alzheimer~\cite{hutton1984eye},
Parkinson~\cite{kuechenmeister1977eye}, or schizophrenia~\cite{holzman1974eye} can be detected using the eye movements as well. Additionally, the eye movements hold information about the activity of the human~\cite{bulling2013eyecontext,steil2015discovery}, the cognitive state~\cite{marshall2007identifying} and personality traits~\cite{10.3389/fnhum.2018.00105}. While all of this information is already critical, several researchers have shown that the gender and age can be also estimated from the eye movements~\cite{cantoni2015gant,sammaknejad2017gender}. While these are useful for applications such as medical diagnosis or security, such information should not be available to everyone.

However, the high and unique information content in the eye tracking signal only becomes clear when biometrics applications are considered. Here, it is possible to unambiguously identify the person by means of the eye behavior. First, approaches required a moving point stimulus which was followed by the user~\cite{kasprowski2004human,kasprowski2004eye,kasprowski2005enhancing} or static images~\cite{maeder2003visual}. Later, users were distinguished using eye movements with a task independent way~\cite{bednarik2005eye}. In addition, model based approaches using gaze behavior with oculomotor models were proposed~\cite{komogortsev2010biometric,komogortsev2013biometric}. Furthermore, distinguishing users while performing different tasks~\cite{eberz2016looks} and a user authentication approach in virtual reality headsets~\cite{zhang2018continuous} were studied using eye movements.

These works show the potential threat to a human by revealing the gaze data. It also means that raw eye tracking data should be handled carefully, especially for storage and transmission purposes. However, there are not many works focusing on privacy-preserving eye tracking. An approach for head mounted eye trackers to detect privacy sensitive situations and to disable eye tracker first person camera using a mechanical shutter was proposed in~\cite{steil2019privaceye}. Privacy-preserving gaze estimation using a randomized encoding based framework and replacing the iris textures of the eye images using rubber sheet model were studied in~\cite{10.1145/3379156.3391364,10.1145/3379156.3391375}, respectively. However, when the personal information protection is taken into account, differential privacy~\cite{dwork2006} provides privacy with theoretical guarantees by adding randomly generated noise. While differential privacy guarantees that adversaries cannot infer whether an individual participated in a database, it also decreases the data utility due to the added noise. The privacy-utility trade-off is usually tailored around a specific use case~\cite{pyrgelis2017knock}, which can be understood as a classification target in the eye tracking world. Recently, differential privacy was applied to eye movements~\cite{steil2019privacy,bozkir2020differential} and heatmaps~\cite{liu2019differential} to protect privacy. Differential privacy is vulnerable to temporal correlations in the data and high dimensionality which are also validated by the recent work. As eye tracking data usually contains long recordings and it is high dimensional, it is challenging to provide the privacy while keeping the utility high. Our approach does not have exactly the same goal with differential privacy as we define a more relaxed version of privacy for eye tracking. With our approach, it is possible to specify the sensitive information that should be hidden in the data, which cannot be achieved with differential privacy.

\subsection{Reinforcement learning}
Reinforcement learning in the area of machine learning refers to one or more agents trying to learn a strategy that maximizes their reward~\cite{kaelbling1996reinforcement,kober2013reinforcement}. The agent in this scenario has different actions that it can perform and after each action it receives a certain reward. For this, different cases have to be considered. The first case are temporal actions similar to a walk through a labyrinth where the agent receives his reward after it tried to go through the labyrinth~\cite{kaelbling1996reinforcement,kober2013reinforcement}. This means that, after executing several actions, the agent receives the final reward. In the second case, the agent has several possible actions without temporal dependency~\cite{kaelbling1996reinforcement,kober2013reinforcement}. In the following, we only deal with the temporally independent applications, because we also pursue this in this work. 

In order to learn complex strategies, there are basically two approaches; one is model-based where a statistical model is given. This model is formulated as a Markov decision problem and is described by states and transitions that are known in advance. For the training of a model-based approach, a multitude of action selection strategy algorithms have been proposed. The first approach is called the greedy algorithm and usually used together with an optimistic initialization~\cite{kaelbling1996reinforcement,kober2013reinforcement}. The second approach in reinforcement learning is called model free. In this approach, the algorithm learns strategies on how to behave under different circumstances. Therefore, the model is not known in advance, but estimated through exploration. The most famous approach is called the Q-learning algorithm~\cite{luong2019applications}. The Q-learning algorithm learns policies for, possibly, an infinite amount of states, where each state can have different amount of actions. It consists of a learning rate and a table that holds the information gathered, the latter is updated with new observations. New actions are chosen using the same selection algorithms as in the model-based approaches. A disadvantage of the Q-learning algorithm is that it is only applicable if the state and action space is small. Therefore, the deep neural networks are employed to replace the table and output the best action by observing the current state. This is called the Deep Q-Learning algorithm (DQL)~\cite{luong2019applications}. In contrast to the tables, the DQL approach has the disadvantage that the neural networks are nonlinear function approximators that only receive the reward for training. This means that the network may not be stable or even diverge~\cite{mnih2015human}. To solve this issue, multiple approaches have been proposed and combined~\cite{luong2019applications}. The first is called the experience replay mechanism. For this approach, the algorithm initializes a replay memory and the initialization is done using the $\epsilon$-greedy algorithm. Out of this memory, mini batches are selected and used for training. Afterwards, the neural network is used to make new experiences, which are stored in the memory. Therefore, the network can always learn on old and new experiences and is thus, stable to train~\cite{luong2019applications}. The second approach to stabilize the training of the neural network is called fixed target Q-network. For this approach, two neural networks are used. The first one is trained based on the memory and then used to slowly update the second network after a fixed set of steps of the learning process~\cite{luong2019applications}. This is especially helpful if the initial exploration is not sufficient.

\section{Method}
\begin{figure}
	\centering
	\includegraphics[width=0.4\textwidth]{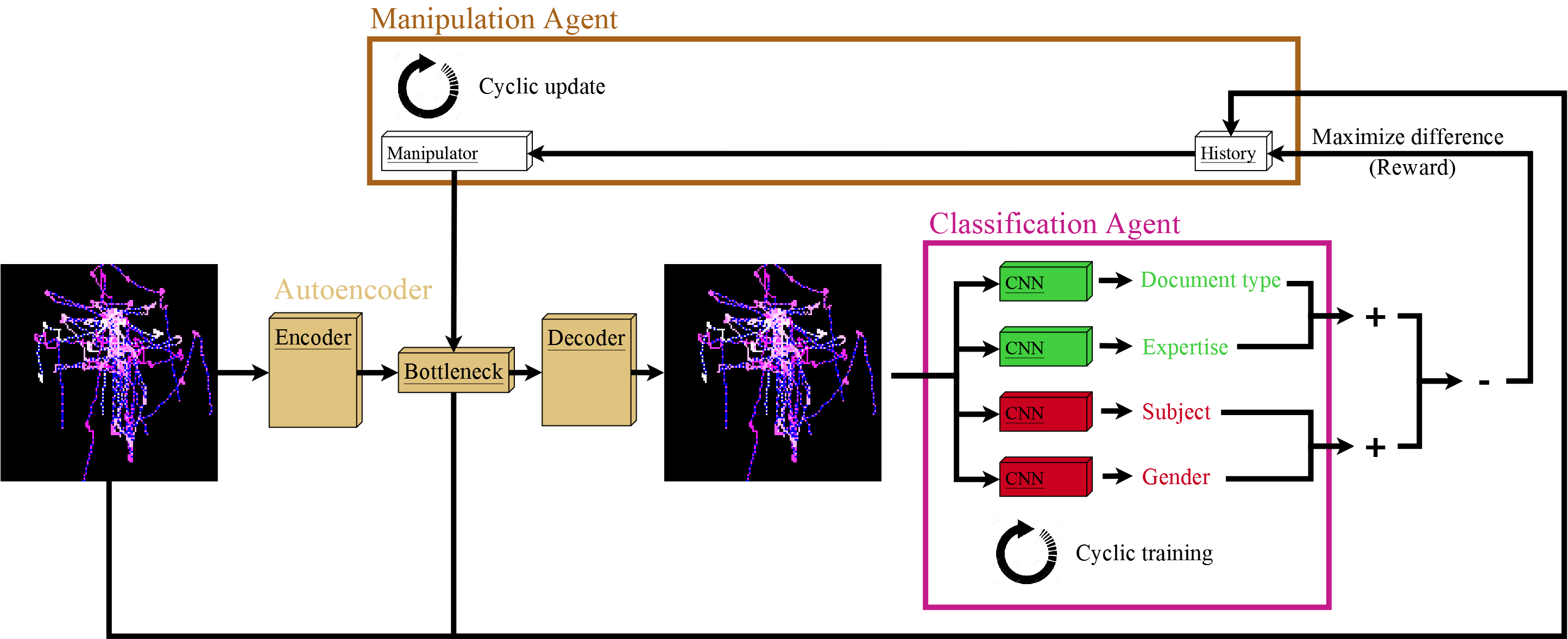}
	\caption{The workflow used for our approach. Classification Agent holds and uses the classifiers and Manipulation Agent the manipulator. Both agents are retrained after a fixed set of steps and have a buffer to hold old and new examples.}
	\label{fig:archall}
\end{figure}
Figure~\ref{fig:archall} shows the general workflow of our approach. The autoencoder is trained preliminary to reconstruct the image. In its central part, it holds values that correspond to general patterns for the reconstruction of the image (Bottleneck in Figure~\ref{fig:archall}). The idea behind using the autoencoder is that it reduces the input data ($64*64*3=12.228$ to $4*4*256=4096$) and, thus, the possible action combinations of the Manipulation Agent as well. Furthermore, in the end it ensures that an image is still generated that is similar to the input image or it consists of general patterns compared to a direct manipulation of the image by the Manipulation Agent. The Manipulation Agent is the reinforcement part of our approach. It learns a manipulation of the bottleneck from the autoencoder based on the previously seen input images and the classification result from Classification Agent. This classification result is only the difference between the good and the bad (Green and red classifiers in Figure~\ref{fig:archall}, respectively) information revealed by the classifiers. The difference is used as a reward in the Manipulation Agent for the performed manipulation, whereas the image itself is the state. The different classification objectives (Document type, expertise, subject, gender) in Figure~\ref{fig:archall} are intended to indicate that our approach supports any number of classifiers. The Manipulation Agent tries to worsen the accuracy of the red classifiers and to keep the accuracy of the green classifiers high. In contrast, the Classification Agent tries to adapt the classifiers to the new image manipulation by retraining them. In the following each part is described in detail.

%%all models in table

\begin{table*}
	\begin{center}
		\caption{The configurations of the used models in our work. The autoencoder is used for extracting high level features from the input image. Classifier A and Classifier B are the networks to classify the subject and stimulus image, respectively. The DQL model is used in the Manipulation Agent as Deep Q-Learning algorithm (DQL).}
		\label{tbl:allmodels}
		\begin{tabular}{c|c|c|c}
			Autoencoder & Classifier A & Classifier B & DQL \\ \hline
			Input $64 \times 64 \times 3$ & $64 \times 64 \times 3$& $64 \times 64 \times 3$& $64 \times 64 \times 3$ \\
			CONV $32 \times 7 \times 7$ & $32 \times 7 \times 7$& $32 \times 7 \times 7$& $32 \times 7 \times 7$ \\
			ReLu, Max Pooling & ReLu, Max Pooling & ReLu, Max Pooling & ReLu, Max Pooling  \\
			CONV $64 \times 7 \times 7$ & $64 \times 7 \times 7$& $64 \times 7 \times 7$& $64 \times 7 \times 7$ \\
			ReLu, Max Pooling & ReLu, Max Pooling & ReLu, Max Pooling & ReLu, Max Pooling  \\
			CONV $128 \times 5 \times 5$ & $128 \times 5 \times 5$ & $128 \times 5 \times 5$ & $128 \times 5 \times 5$ \\
			ReLu, Max Pooling & ReLu, Max Pooling & ReLu, Max Pooling & ReLu, Max Pooling  \\
			CONV $256 \times 5 \times 5$ & $256 \times 5 \times 5$ & $256 \times 5 \times 5$ & Fully $4096$ \\
			ReLu & ReLu, Max Pooling & ReLu, Max Pooling & -  \\
			TCONV $128 \times 5 \times 5$ & Fully $512$ & Fully $512$ & - \\
			ReLu & ReLu & ReLu & -  \\
			TCONV $64 \times 5 \times 5$ & Fully \#Classes & Fully \#Classes & - \\
			ReLu & - & - & -  \\
			TCONV $32 \times 7 \times 7$ & - & - & - \\
			ReLu & - & - & -  \\
			TCONV $3 \times 7 \times 7$ & - & - & - \\ \hline
		\end{tabular}
	\end{center}
\end{table*}

The first column of Table~\ref{tbl:allmodels} shows the architecture of the used autoencoder. Each convolution block is followed by a rectifier linear unit (ReLu) and max pooling for size reduction. For the decoder of the autoencoder, we used transposed convolutions instead of pooling. The input to the network is an image with size $64 \times 64 \times 3$. The bottleneck in the autoencoder is the block with size $4 \times 4 \times 256$. For the training, we used stochastic gradient decent (SGD) with an initial learning rate of $10^{-2}$, decreasing each 200 epochs by a factor of $10^{-1}$. The training stops at a learning rate of $10^{-7}$. Weight decay and momentum were to $5*10^{-4}$ and $9*10^{-1}$, respectively. During the training, we used a batch size of $40$ and the L2 loss formulation. The autoencoder is trained only once before starting our reinforcement learning approach. 

The classifiers in the Classification Agents use a similar structure as the autoencoder and details of Classifier A and B are depicted in Table~\ref{tbl:allmodels} second and third columns, respectively. Each convolution block uses a ReLu together with a max pooling operation. Before the first fully connected layer, we used a dropout, which deactivates $50\%$ randomly. Classifiers A and B in Table~\ref{tbl:allmodels} have the same structure except for the last fully connected layer, which has either eight (Subject) or four (Stimulus image) output neurons. For the training, we used SGD with an initial learning rate of $10^{-4}$ decreasing each 500 epochs by a factor of $10^{-1}$. The training stops at a learning rate of $10^{-7}$. Weight decay and momentum were set to $5*10^{-4}$ and $9*10^{-1}$, respectively. During training, we used a batch size of $50$ and the log multi-class loss with softmax.

Since these classifiers are subject to the cyclic training of the Classification Agent, they are always re-trained once the reinforcement learning has been stabilized. This new training is done with random initialization. The idea behind is that the convolutions, which learn new feature extractors, adapt to the new image manipulation and, thus, improve the classification result. The training itself is done using all the manipulated images seen so far in addition to the non-manipulated ones (only from the training set).

\begin{figure}[h]
	\centering
	\includegraphics[width=0.4\textwidth]{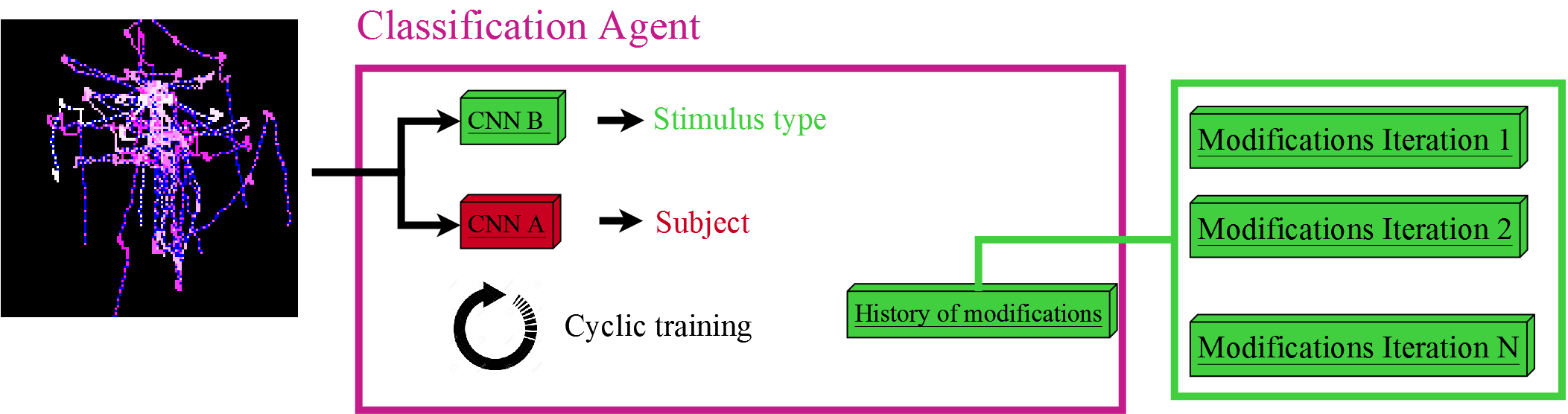}
	\caption{The setup of the Classification Agent with a memory for manipulated data seen in the past.}
	\label{fig:agent1}
\end{figure}

We show the workflow for the Classification Agent with the memory in Figure~\ref{fig:agent1}. In comparison to Figure~\ref{fig:archall}, which is a general overview, it can be seen that we now have only two classes. Those two classes are also used in our experiments for the evaluation section which is why we decided to insert them in the detailed view of the Classification Agent. In the memory (Figure~\ref{fig:agent1}) are all the seen manipulated images from the training set together with their labels. Images from the validation set are discarded and, therefore, not stored in the memory of the Classification Agent. For the training and test sets, we made a $50\%$ to $50\%$ split. We separated the data to produce an equal amount of stimulus and subject classes. As it can be seen, Classification Agent does not use reinforcement learning. This agent can be understood as a supervised learner, which retrains its classifiers.

In contrast to the Classification Agent, the Manipulation Agent uses reinforcement learning for training. The used DQL model is shown in the fourth column of the Table~\ref{tbl:allmodels}. It consists of three convolution blocks and a fully connected output layer. The input of this model is the current image, which is called the state, and the output ($4096$ fully connected neurons) are the actions. Between each convolution block, we used ReLu and max pooling as in the previous models. The output of the last layer was set to $1$ if it was greater or equal to $0.5$. Otherwise it was set to $0$, meaning that our model could either deactivate a feature in the bottleneck of the autoencoder or let it unchanged. For the training, we used SGD with a fixed learning rate of $10^{-4}$. The training stops after $10$ epochs of training on the entire memory of the Manipulation Agent. Weight decay and momentum were set to $1*10^{-5}$ and $9*10^{-1}$, respectively. During the training, we used a batch size of $100$ and the L2 loss formulation for reinforcement learning $(predicted - actual)^2$. The parameter $predicted$ in this context means the result of DQL1 from the current input image. Since there is no ground truth in reinforcement learning, the parameter $actual$ is computed based on a second network (DQL2) and the reward $R$. Therefore, the ground truth is formulated as $actual = R + y * DQL2$. As mentioned before, $R$ is the reward (Result of Classification Agent), $DQL2$ is the output of a second network and $y$ is the discount factor, which is adjusted through training so that the network explores more in the beginning. This usage of two neural networks is called a fixed target Q-network~\cite{luong2019applications}. Therefore, after $10$ training runs of DQL1, we set $DQL2=DQL1$ since DQL1 had stabilized.

\begin{figure}[h]
	\centering
	\includegraphics[width=0.4\textwidth]{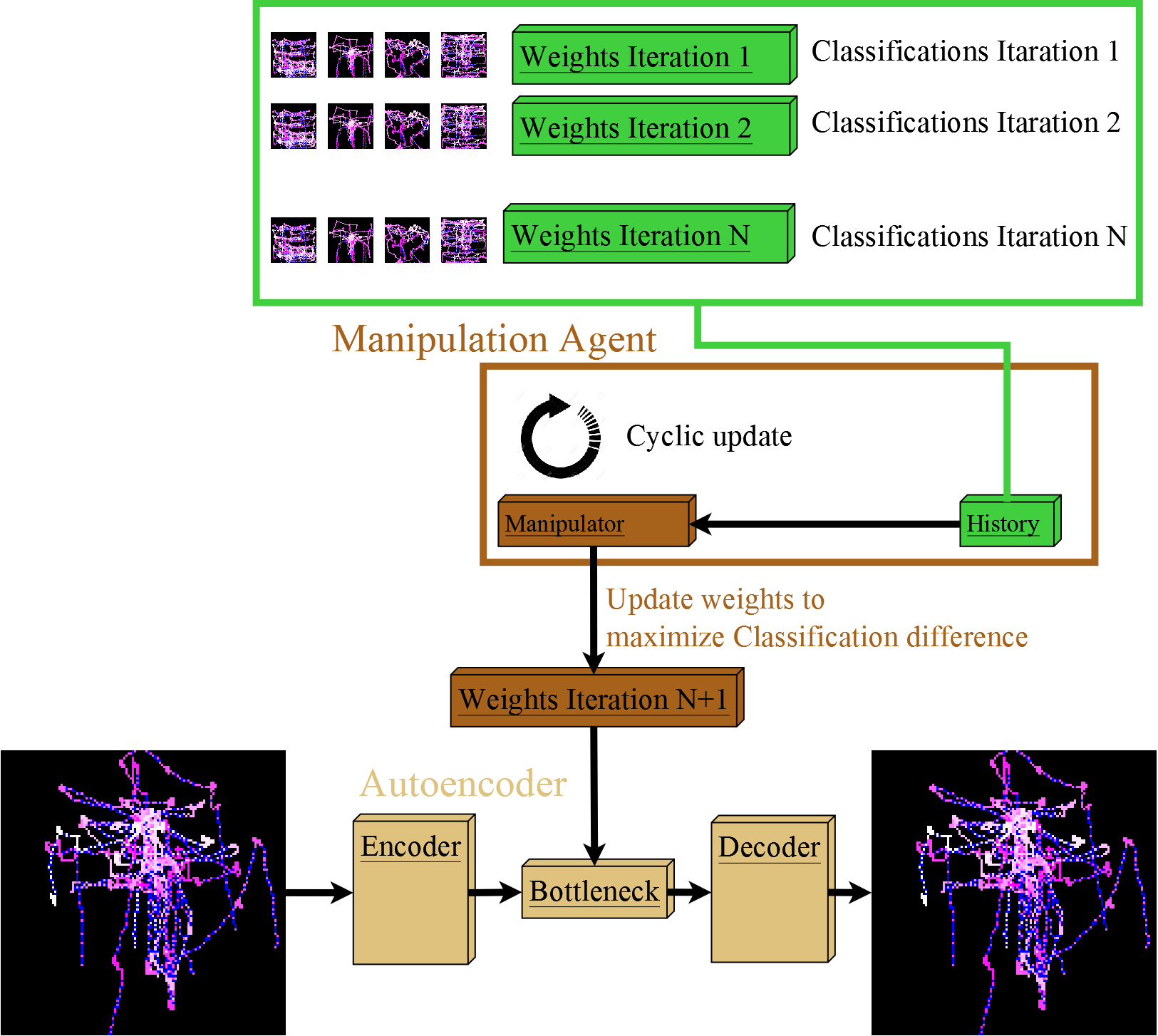}
	\caption{The memory and setup of the Manipulation Agent.}
	\label{fig:agent2}
\end{figure}

In addition to the fixed target network, we use the experience replay mechanism~\cite{luong2019applications} as can be seen in Figure~\ref{fig:agent2}. As mentioned in the related work, this concept describes the memory which holds all examples (Stimulus, actions, and classification result). In the memory, we only store examples from the training set, since we want to evaluate our approach especially for unseen data. The memory is initialized before starting the entire approach and the networks DQL1 and DQL2 are trained on it. For this initialization, we compute the change of each value in the bottleneck on the classification and store it in the memory of the Manipulation Agent. In addition, we compute $100$ random changes of 2-100 values in the bottleneck. This means that for the change of two values, we compute $100$ random changes and the same for three values, four values, and so on.

For data augmentation of all models, we used random noise which was in the range of $0$-$20\%$, cropping and shifting the scanpath. Cropping means the extraction of the $60$-$100\%$ of the scanpath randomly and drawing it on the input image. Shifting means randomly selected constant shift of the entire scanpath, where we selected in the range of $0$-$30\%$ of the stimulus size.

In addition to our reinforcement learning approach, we have evaluated differential privacy particularly in terms of utility and a Generative Adversarial Network (GAN) namely a supervised approach to justify the usage of reinforcement learning for the manipulation. While our approach does not provide formal privacy guarantees as differential privacy and we provide more relaxed version of privacy while keeping the utility high, differential privacy is also vulnerable, especially to the correlated and high dimensional data. Therefore, differential privacy, in our context, provides less privacy compared to applications in the domain of databases~\cite{8269219}. For evaluation, we opted for the standard Laplacian mechanism of the $\epsilon$-Differential Privacy ($\epsilon$-DP) applied both on raw eye tracking data (DP-Raw) and generated image (DP-Image)~\cite{dwork2006,sarathy2011evaluating}. In the differential privacy, the amount of added noise is generated using function sensitivities ($\Delta f$) and an $\epsilon$ parameter. For the function sensitivities, we used $L_1$ sensitivities which are calculated as the maximum Manhattan distance between recordings~\cite{rastogi2010differentially} and maximum pixel distance per each channel of the images for DP-Raw and DP-Image, respectively. $N$-sized randomly generated Laplacian noise vectors are calculated as $Lap^N(\lambda) = Lap^N(\Delta f/ {\epsilon})$, where $N$ denotes size of the noise added data. For the evaluation, we applied the Laplacian mechanism $100$ times and averaged the results accordingly. We used majority voting while selecting the detected class by the networks. To find the optimal utility-privacy ratio, we evaluated various $\epsilon$ values in the search range of $[0.01-15.0]$ and $[10.00-500.0]$ by $0.01$ sized steps for the images and raw gaze data, respectively. For the images, the $\epsilon$ values are multiplied by the image resolution ($64 \times 64$) as the differential privacy is preserved by the Sequential Composition Theorem due to the independency of the image pixels~\cite{mcsherry2009privacy}. Therefore, the search range for the $\epsilon$ was $[40.96-61440]$, but the $0.01$ search steps are made based on the single pixel search range ($[0.01-15.0]$). We skipped $\epsilon$ values for the raw eye tracking data if there were less than three gaze points remaining on the image. The optimal $\epsilon$ was selected based on the maximum distance between the stimulus and subject classification, where the subject classification was at chance level. 

In addition, we have evaluated a supervised learning approach to justify the usage of reinforcement learning. The same models as shown in Table ~\ref{tbl:allmodels} were used and trained as a GAN. The autoencoder is used as the generator, whereas the Classifiers A and B are used as discriminators. Before we trained the GAN, we initially trained the Autoencoder, Classifier A, and Classifier B for $100$ epochs with the already provided training parameters. This was done to stabilize the training of the GAN afterwards. To adapt the initial training to the training of the GAN, we added the logarithmic loss from the generated image as was done in~\cite{goodfellow2014generative} with the difference that the classifiers still had to predict the correct class.

For the generator (G), we used the formulation of ($log(1-D(G(I)))$)~\cite{goodfellow2014generative} but in our case the discriminator (D) consists of two networks. Therefore, Classifier A and Classifier B can only contribute $0.5$ each but in inverse directions. This means that if Classifier A is correct, it contributes $0.5$ and if Classifier B is wrong then it additionally contributes $0.5$ since we want the GAN to learn to preserve the information classified by the Classifier A and hide the information important for the Classifier B. Based on the softmax output, we simply compute the probability for the correct class for the Classifiers A and B and weight them both with $0.5$.

\section{Evaluation}
In this section, we give an overview on the used datasets and discuss our results.
\subsection{Datasets}
\begin{figure}
	\centering
	\includegraphics[width=0.4\textwidth]{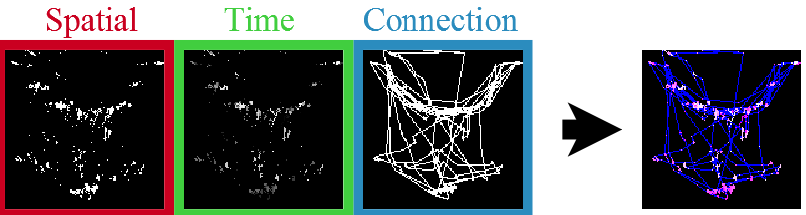}
	\caption{The encoding of eye tracking data as image.}
	\label{fig:dataman}
\end{figure}

\textbf{ETRA 2019 Challenge dataset ~\cite{otero2008saccades,mccamy2014highly}:} A dataset with $8$ subjects and $120$ trials per subject. Therefore, it consists of $960$ trials with a length of $45$ seconds per trial. The dataset includes different tasks, namely, visual fixation, visual search, and visual exploration. Additionally, four different stimuli were presented; which are blank, natural, where is waldo, and picture puzzle. For the image generation, out of the raw gaze data files, we used the approach from \cite{emoji2019fuhl}. This means that the raw gaze data is in the red channel as dots, the green channel holds the time by adjusting the intensity of the dot, and the blue channel holds the relationship of the gaze points by connecting them as lines, which can be seen in Figure~\ref{fig:dataman}.

%http://webmail.inb.uni-luebeck.de/inb-toolsdemos/EyeMovementDataSet.html
\textbf{Gaze~\cite{dorr2010variability}:} A data set with eye tracking data on dynamic scenes. The data was recorded using an SR Research EyeLink II eye tracker with $250$ Hz. For our experiment, we used the data provided for static images where each static image of a video was considered the same image. In addition, we excluded subject V01 since there was only one recording available. Therefore, we used the eye tracking data of $10$ subjects on $9$ images for our experiment with an average recording length of $2$ seconds. The training and test split was done using $50\%$ for the training and $50\%$ for the validation with a random selection. To treat both classifiers equally, the training set contained data from each subject and image.

%http://people.csail.mit.edu/tjudd/WherePeopleLook/index.html
\textbf{WherePeopleLook~\cite{Judd2009} (WPL):} An eye tracking dataset for integrating top-down features into saliency map generation. It consists of 1003 static images with eye tracking data of 15 subjects per image with an average recording length of $3$ seconds. For our experiment, we used a $50\%$-$50\%$ split where the training data included all subjects and images at least once to treat both classifiers equally.

%http://live.ece.utexas.edu/research/doves/
\textbf{DOVES~\cite{bovik2009doves}:} An eye tracking dataset of $29$ subjects on $101$ natural images with an average recording length of $5$ seconds. The recordings were performed with a $200$ Hz high-precision dual-Purkinje eye tracker. Similar to the WherePeopleLook~\cite{Judd2009} dataset, we made a $50\%$-$50\%$ training and test split. The training data included each subject and image at least once to treat both classifiers equally.

\subsection{Results}
For our first two experiments, we used the ETRA 2019 Challenge dataset. The first experiment shows the results of our approach for different iterations, as well as before and after the adaption of the classifiers (Classification Agent). This experiment shows that our approach is capable of removing unwanted information in the scanpath. In this scenario, it is the information of the subject. Table~\ref{tbl:classificationresults} shows the classification results per iteration. With iteration, we mean that the reinforcement learning (RL), namely the Manipulation Agent has stabilized, which are approximately $1000$ training runs. After each iteration, the Classification Agent starts to retrain the classifiers, which is indicated by the adaption rows. RL-Initial in Table~\ref{tbl:classificationresults} corresponds to the initial results of the pretrained classifiers. The chance level is shown at the bottom of Table~\ref{tbl:classificationresults}. As can be seen, the Manipulation Agent always succeeds in dropping the classification accuracy for the subject close to the chance level. Afterwards, the Classification Agent adapts the classifiers, but with less success for the subject classification if the process over all iterations is considered. In the last iteration ($20$), the training of the subject classifier fails and is close to the chance level. This is also the case for DP-Raw (Differential privacy applied to the raw eye tracking data), DP-Image (Differential privacy applied to the image), and the GAN (Generative Adversarial Network) approach. It is clear that our reinforcement learning approach performs better than differential privacy in terms of receiving the stimulus information, namely the utility. In addition, our approach performs slightly better than the GAN approach.
\begin{table}
	\begin{center}
		\caption{Accuracy of the classifiers after each iteration and before as well as after the adaption of Classification Agent. RL is the proposed approach, GAN is the same models trained supervised, DP-Raw is the Differential Privacy applied to the raw gaze data, and DP-Image is the differential privacy applied to the image. The best results are in bold.}
		\label{tbl:classificationresults}
		\begin{tabular}{ccccc}
			& \multicolumn{2}{c}{No Adaption} & \multicolumn{2}{c}{Adaption} \\
			Iteration & Stim & Sub & Stim & Sub \\ \hline
			RL-Initial & - & - & $0.96$ & $0.93$ \\
			RL-$1$ & $0.95$ & $0.13$ & $0.96$ & $0.93$ \\
			RL-$2$ & $0.95$ & $0.11$ & $0.95$ & $0.91$ \\
			RL-$5$ & $0.91$ & $0.12$ & $0.93$ & $0.52$ \\
			RL-$10$ & $0.88$ & $0.14$ & $0.91$ & $0.31$ \\
			RL-$15$ & $0.78$ & $0.12$ & $0.86$ & $0.22$ \\
			RL-$20$ & $0.81$ & $0.13$ & $\mathbf{0.83}$ & $\mathbf{0.15}$ \\ \hline
			GAN & $0.75$ & $0.13$ & $0.81$ & $\mathbf{0.15}$ \\ \hline
			DP-Raw $\epsilon = 238.83$& $0.27$ & $0.15$ & $0.30$ & $\mathbf{0.15}$ \\  \hline
			DP-Image $\epsilon = 10485.76$ & $0.41$ & $0.11$ & $0.59$ & $\mathbf{0.15}$ \\  \hline
			Chance level & $0.25$ & $0.12$ & $0.25$ & $0.12$ \\ \hline
		\end{tabular}
	\end{center}
\end{table}

In the second experiment, we evaluate the importance of different channels of the input image for different iterations of our approach. This experiment shows the advantage of our approach to other privacy preservation methods since the feature extractors (Neural networks in the Classification Agent) adapt to the new image manipulation as well as our image manipulation technique. For all experiments, we used a $50\%$ split of the data where the test and validation set contain always equal amount of subjects and stimuli samples. Table~\ref{tbl:importamnce} shows the percentage amount of changed values per channel normalized over the total amount of changed values. Due to the construction of the image with raw dots in the red channel, connected dots in the blue channel, and the time as intensity value per dot in the green channel, we estimated the importance of their contribution. For iteration $1$, it can be seen that the subject information was mainly extracted out of the red channel, which holds only spatial information. In the second iteration, this swaps to the blue channel, which holds the interconnections between the gaze points and, therefore, the spatial information. After $5$ iterations, the amount of changes have balanced per channel. If we compare these results to Table~\ref{tbl:classificationresults}, it can be seen that it had already a significant impact on the adaption of the subject classifier. After the last iteration, the amount of changes has again nearly balanced, where the green channel is the lowest. Since the green channel is the only channel that has temporal information, it could be argued that it is less important for the subject information since the green channel also contains spatial information. This statement is purely hypothetical and requires further experiments and research as well as another construction of input data.

\begin{table}
	\begin{center}
		\caption{Importance of spatial (R, B; red and blue channel) and temporal (G; intensity in green channel) features for the classification per iteration. The importance is measured in percentage of values changed in total per channel.}
		\label{tbl:importamnce}
		\begin{tabular}{cccc}
			Iteration & Red (Spat.) & Green (Temp.) & Blue (Spat.) \\ \hline
			$1$ & $72\%$ & $16\%$ & $12\%$ \\
			$2$ & $19\%$ & $12\%$ & $69\%$ \\
			$5$ & $36\%$ & $33\%$ & $31\%$ \\
			$10$ & $41\%$ & $20\%$ & $39\%$ \\
			$15$ & $38\%$ & $22\%$ & $40\%$ \\
			$20$ & $37\%$ & $28\%$ & $35\%$ \\ \hline
		\end{tabular}
	\end{center}
\end{table}

In the third experiment, we use the data manipulation (learned with reinforcement learning) and the autoencoder on other public data sets without further training. However, the classifiers are re-trained on the output of the autoencoder and additionally adapted to the data manipulation in the further step. For the classification on the datasets Gaze, WherePeopleLook, and DOVES, we used the same model as in Experiment $1$ (Table~\ref{tbl:allmodels}). For training, we set the initial learning rate to $10^{-2}$ and reduced it by a factor of $10^{-1}$ every 100 epochs until we reached $10^{-7}$. The optimizer used was SGD with weight decay of $5*10^{-4}$ and momentum of $0.9$. For the dataset Gaze, we used a batch size of twice the number of classes and made sure that there were always $2$ examples of each class in a batch. For the WherePeopleLook, we used double the number of classes for the subject classification as well. For the Stimulus Classification, we used only the single class number as batch size. For the DOVES dataset, we used twice the number of classes as batch size for both classifiers as for the Gaze dataset and made sure that there were always two examples of each class per batch as well.
\begin{table}[h]
	\setlength\tabcolsep{0.05em}
	\begin{center}
		\caption{Accuracy on new unseen datasets with retrained classifiers but the same data manipulation learned from experiment $1$ and $2$ as well as the same weights for the autoencoder. RL is the proposed approach, GAN is the same models trained supervised, DP-Raw is the differential privacy applied to the raw gaze data, and DP-Image is the differential privacy applied to the image. The best results are in bold.}
		\label{tbl:newunseen}
		\begin{tabular}{cc|cc|cc|cc}
			&& \multicolumn{2}{c|}{None} & \multicolumn{2}{c|}{Manipulation}  & \multicolumn{2}{c}{Adapted}\\
			Data & Method & Stim & Sub & Stim & Sub & Stim & Sub \\ \hline
			\multirow{6}{*}{\rotatebox{90}{Gaze}} & RL & \multirow{6}{*}{$75$} & \multirow{6}{*}{$31.66$} & $40$ & $8.88$ & $\mathbf{71.11}$ & $\mathbf{13.33}$\\
			& GAN &  &  & $37.24$ & $14.32$ & $61.64$ & $19.53$\\  
			& DP-Raw  &  &  & \multirow{2}{*}{$15.44$} & \multirow{2}{*}{$12.54$} & \multirow{2}{*}{$16.25$} & \multirow{2}{*}{$13.96$}\\  
			& $\epsilon = 32.85$ &  &  &  &  &  & \\   
			& DP-Image &  &  & \multirow{2}{*}{$21.22$} & \multirow{2}{*}{$11.81$} & \multirow{2}{*}{$59.83$} & \multirow{2}{*}{$13.61$} \\
			& $\epsilon = 34938.88$ &  &  &  &  &  &  \\
			& Chance & $11.11$ & $10$ & $11.11$ & $10$  & $11.11$ & $10$ \\ \hline
			\multirow{6}{*}{\rotatebox{90}{WPL}} & RL & \multirow{6}{*}{$31.23$} & \multirow{6}{*}{$30.06$} & $21.54$ & $6.39$ & $\mathbf{30.48}$ & $\mathbf{8.28}$  \\
			& GAN &  &  & $18.47$ & $14.76$ & $26.74$ & $20.37$\\  
			& DP-Raw & &  & \multirow{2}{*}{$0.19$} & \multirow{2}{*}{$7.09$} & \multirow{2}{*}{$0.4$} & \multirow{2}{*}{$8.93$}\\  
			& $\epsilon = 73.66$ & &  &  &  &  & \\  
			& DP-Image &  &  & \multirow{2}{*}{$7.15$} & \multirow{2}{*}{$6.72$} & \multirow{2}{*}{$8.41$} & \multirow{2}{*}{$8.91$}\\ 
			& $\epsilon = 16547.84$ &  &  &  & &  & \\ 
			& Chance & $0.099$ & $6.66$ & $0.099$ & $6.66$  & $0.099$ & $6.66$ \\ \hline
			\multirow{6}{*}{\rotatebox{90}{DOVES}} & RL & \multirow{6}{*}{$10.86$} & \multirow{6}{*}{$44.90$} & $4.3$ & $6.69$ & $\mathbf{9.15}$ & $13.66$  \\
			& GAN &  &  & $5.68$ & $6.73$ & $8.26$ & $19.55$\\  
			& DP-Raw &  &  & \multirow{2}{*}{$1.81$} & \multirow{2}{*}{$3.95$} & \multirow{2}{*}{$1.42$} & \multirow{2}{*}{$\mathbf{12.45}$}\\ 
			& $\epsilon = 438.15$ &  &  &  &  &  & \\ 
			& DP-Image &  &  & \multirow{2}{*}{$1.14$} & \multirow{2}{*}{$5.01$} & \multirow{2}{*}{$1.5$} & \multirow{2}{*}{$22.11$}\\ 
			& $\epsilon = 13762.56$ &  &  &  &  &  & \\ 
			& Chance & $0.99$ & $3.44$ & $0.99$ & $3.44$ & $0.99$ & $3.44$ \\ \hline
		\end{tabular}
	\end{center}
\end{table}

Table~\ref{tbl:newunseen} shows the results of the third experiment. The results without the data manipulation are depicted in the first column. Comparing these with the results on the Challenge dataset in Table~\ref{tbl:classificationresults}, it is seen that the results are significantly lower. One of the reasons is that there are more number of classes which increases the challenge for the classification, but the main reason is the significantly lower recording time. For the Challenge data set, the average recording time is $45$ seconds.  In comparison, Gaze, WherePeopleLook, and DOVES datasets have an average of $2$, $3$, and $5$ seconds, respectively. This shows that the Challenge dataset provides a multiple of the information for the neural networks. It means that the data from the Challenge dataset contains significantly more personal information as well as more information about the structure of the stimuli. It is also interesting how little eye tracking data is sufficient to classify a subject. For instance, if the results of the DOVES dataset are compared with Gaze and WherePeopleLook datasets, it is seen in the first column of Table~\ref{tbl:newunseen} that DOVES has a higher accuracy for the subject classification although it has a lower chance level, but a $2$-$3$ seconds longer recording time. In contrast, the detection rate for the stimuli classification is significantly lower compared to the other datasets.

The results after the data manipulation by the Manipulation Agent are shown in the second column of Table~\ref{tbl:newunseen}. The Manipulation Agent has not been retrained and neither has the autoencoder. As it is shown in the results, the data manipulation has a significant impact on the accuracy of the classifiers. It holds for the stimulus and subjects, although the subject classification is influenced more, except for the DOVES dataset as everything is reduced below the chance level. Since this can be purely due to data augmentation, we have also adapted the classifiers to the data manipulation via training. For this purpose, the training examples were manipulated with the Manipulation Agent and both the unaltered and the manipulated data were used for the training. The results are be shown in the third column of Table~\ref{tbl:newunseen}. While the subject classification in the DOVES dataset is still significantly above the chance level with $13.66\%$, the personal information was mainly removed in the other two datasets. The stimulus information was mainly retained for all datasets which empirically shows that it is possible to find generalized patterns to hide specific information using our approach.

\section{Conclusion}
In this work, we showed the applicability of reinforcement learning for removing personal information from eye tracking data. In addition, it can be used to evaluate the features and is able to adapt to an adaptive attacker. Our approach is able to remove and preserve information of multiple classification targets. We empirically showed that our approach has generalized and is also applicable to unseen data sets. This is interesting since it could mean that our approach can be applied to improve the robustness of neural networks as a pre-processing module or during training as an adversarial attack generator.

\bibliography{egbib}

\end{document}